\pdfoutput=1
\documentclass[11pt]{article}

\usepackage[]{ACL2023}

\usepackage{times}
\usepackage{latexsym}

\usepackage[T1]{fontenc}
\usepackage[utf8]{inputenc}

\usepackage{microtype}
\usepackage{graphicx}
\usepackage{inconsolata}

\title{Improving Multimodal Contrastive Learning of Sentence Embeddings with \\ Object-Phrase Alignment}

\author{Kaiyan Zhao, Zhongtao Miao, Yoshimasa Tsuruoka\\
        The University of Tokyo, Tokyo, Japan \\
        \texttt{\{zhaokaiyan1006, mzt, yoshimasa-tsuruoka\}@g.ecc.u-tokyo.ac.jp}}

\begin{document}
\maketitle
\begin{abstract}
Multimodal sentence embedding models typically leverage image-caption pairs in addition to textual data during training. However, such pairs often contain noise, including redundant or irrelevant information on either the image or caption side. To mitigate this issue, we propose MCSEO, a method that enhances multimodal sentence embeddings by incorporating fine-grained object-phrase alignment alongside traditional image-caption alignment. Specifically, MCSEO utilizes existing segmentation and object detection models to extract accurate object-phrase pairs, which are then used to optimize a contrastive learning objective tailored to object-phrase correspondence. Experimental results on semantic textual similarity~(STS) tasks across different backbone models demonstrate that MCSEO consistently outperforms strong baselines, highlighting the significance of precise object-phrase alignment in multimodal representation learning.\footnote{Work in progress.}
\end{abstract}

\section{Introduction}
Sentence embedding aims to capture the semantic meaning of sentences by projecting them into a fixed vector~(embedding) in a shared high-dimensional space~\cite{reimers-gurevych-2019-sentencebert, gao-etal-2021-simcse}. The mainstream for training sentence embedding models lies in contrastive learning~(CL), where the distances between positive examples are pulled closer and those among negative examples are pushed farther~\cite{oord2019representationlearningcontrastivepredictive}.  

Recently, some works have been trying to incorporate non-linguistic information, such as visual context, alongside textual information to improve sentence embedding models~\cite{zhang-etal-2022-mcse, non-lingustic-supervision, visually-supervised-pretraining, nguyen-etal-2024-kdmcse}. The resulting representations are referred to as \textit{multimodal sentence embeddings}, most of which are trained using image-caption pairs to align visual and textual modalities.

As discussed in ~\citet{nguyen-etal-2024-kdmcse}, current image-caption datasets often contain some noise, which will affect the construction of negative samples for contrastive learning. For instance, if two visually similar images, both of which could plausibly match the same caption, are treated as negative pairs, the training signal becomes misleading. Building on this observation, we further argue that even image-caption pairs labeled as positive samples can contain significant noise.

\begin{figure}[t]
    \centering
    \includegraphics[width=1.0\linewidth]{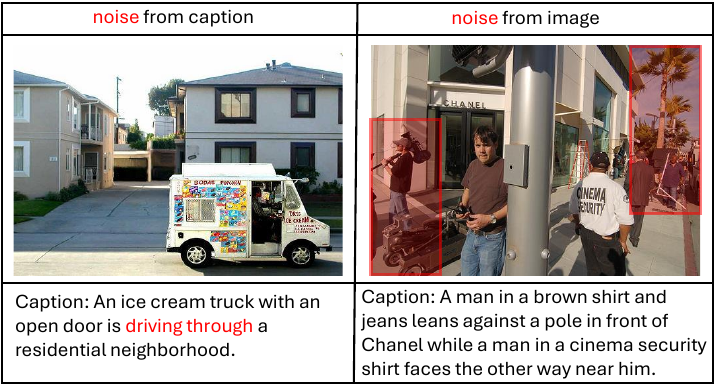}
    \caption{Image-caption pair examples from the Flickr30k dataset. We use red color to highlight the potential sources of noise. \textbf{Left}: The caption includes ambiguous or inaccurate details not supported by the image content. \textbf{Right}: The image contains complex visual elements that are not described in the caption.}
    \label{fig:intro}
\end{figure}

Specifically, we present two image-caption pair examples in Figure~\ref{fig:intro} from the Flickr30k dataset~\cite{flickr} to illustrate this problem. We mainly focus on two types of noise: from the caption side and from the image side. In the left example, the phrase ``driving through'' in the caption does not accurately reflect the visual content, as the car in the image appears to be stopped with its door open. In the right example, the image contains additional elements, such as the cameraman and background trees, that are not mentioned in the caption, leading to incomplete alignment between modalities.

\begin{figure*}[h]
\centering
\includegraphics[width=1.0\textwidth]{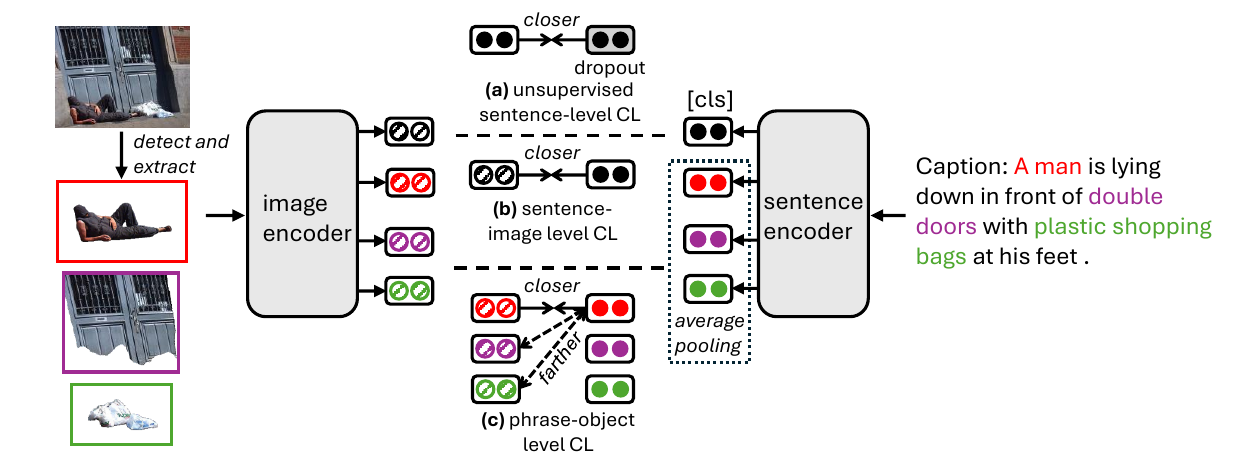}
\caption{Overview of MCSEO. Given an image-caption pair, we first decompose the pair into object-phrase pairs and then apply three training objectives.}
\label{method}
\end{figure*}

Since these image-caption pairs are commonly used as positive examples during the training of multimodal sentence embeddings, we argue that directly minimizing the distance between noisy image-caption pairs may degrade model performance. While we acknowledge that constructing perfectly aligned image–caption pairs is inherently challenging, this work aims to address the noise present in existing datasets by introducing finer-grained alignment: not only between images and captions, but also between objects and phrases. To this end, we propose MCSEO (Multimodal Contrastive Learning of Sentence Embeddings with Object-Phrase-Level Alignment), a framework that leverages existing segmentation and object detection models to extract objects from images and selectively decompose captions into phrases corresponding to these visual elements. This allows us to obtain more accurate object–phrase pairs with reduced noise. We further introduce an object-phrase-level contrastive learning objective to help the model better capture this fine-grained alignment.

Experimental results on seven STS tasks across multiple backbone models demonstrate the effectiveness of MCSEO, which consistently outperforms existing multimodal sentence embedding baselines. We attribute this improvement to the use of accurate object-phrase pairs, which provide more precise alignment signals during training and lead to better sentence representations.

\section{Related Works}
Sentence embedding models have shown strong performance under contrastive learning objectives, where semantically similar sentences are pulled closer in embedding space, and dissimilar ones are pushed apart~\cite{reimers-gurevych-2019-sentencebert, gao-etal-2021-simcse}.

Recently, many works have explored multimodal sentence embeddings, which incorporate non-linguistic information from other modalities such as images~\cite{zhang-etal-2022-mcse, non-lingustic-supervision, visually-supervised-pretraining, nguyen-etal-2024-kdmcse}.

MCSE~\cite{zhang-etal-2022-mcse} makes an early attempt to enhance sentence embeddings by pulling image and caption embeddings closer together using image–caption datasets. Their method employs an image encoder to obtain visual embeddings, which are then aligned with corresponding caption embeddings through contrastive learning besides traditional textual datasets. KDMCSE~\cite{nguyen-etal-2024-kdmcse} further investigates the issue of noise in image–caption datasets, particularly in the construction of negative samples, highlighting that visually distinct images may sometimes be described with identical or highly similar captions.

While KDMCSE focuses on reducing the impact of noisy negatives, our work addresses the complementary problem: improving the quality of positive pairs. Specifically, we propose to construct more reliable positive examples by aligning fine-grained elements within image–caption pairs, such as object–phrase correspondences, thereby reducing the noise inherent in standard multimodal supervision.

\section{Methods}
An overview of our proposed MCSEO framework is illustrated in Figure~\ref{method}. MCSEO employs three contrastive learning objectives during training: (a) unsupervised sentence-level contrastive learning, (b) sentence–image level contrastive learning, and (c) our novel object–phrase level contrastive learning. We first describe objectives (a) and (b), followed by a detailed explanation of how we extract object–phrase pairs and incorporate the object–phrase level contrastive learning objective in MCSEO.

\subsection{Unsupervised Sentence-level CL}
The purpose of sentence embedding is to train a model $f_\theta$ that transforms a text $x_i$ into a fixed-size vector $h_i=f_\theta(x_i)$, where $x_i$ is a sequence from dataset $\mathcal{D}_{text}$. 
In unsupervised CL for sentence embeddings, dropout is often used as the positive example $x_i^+$ for $x_i$, while other sequences from the training batch serve as negative examples. The training objective can be shown in the following equation~\cite{gao-etal-2021-simcse}:
\begin{equation}
   \\ l_{i}^{text} = -\mathrm{log}\frac{e^{sim(h_i, h_i^+)/\tau}}{\sum_{j=1}^{N} e^{sim(h_i, h_j)/\tau}},
\end{equation}
where $sim()$ is a metric for calculating similarity, and $\tau$ is the temperature parameter. $N$ is the size of a mini-batch. In this way, the model is trained to pull the distance between $h_i$ and $h_i^+$ close and push the distance between $h_i$ and other $h_j$ farther.

\begin{table*}[ht]
\centering
\resizebox{0.9\textwidth}{!}{
\begin{tabular}{lcccccccc}
\hline
Model                    & STS12 & STS13                & STS14                & STS15                & STS16                & STS-B & SICK-R & avg. \\ \hline \hline
\multicolumn{9}{c}{\textit{text-only models}}                                                                                                                        \\ \hline
BERT                     &   39.7 &  59.4 &  49.7 & 66.0 & 66.2 & 53.9 & 62.1 & 56.7       \\ 
RoBERTa  &    40.9   &  58.7   & 49.1  & 65.6  & 61.5  & 58.6  & 61.6  & 56.6      \\ 
SimCSE-BERT  &   68.4  & 82.4   & 74.4   & 80.9   &78.6    &76.9   &72.2    &76.3    \\ 
SimCSE-RoBERTa & 70.2 & 81.8 & 73.2 & 81.4 & 80.7 & 80.2 & 68.6 & 76.6 \\

\hline
\multicolumn{9}{c}{\textit{text-image models}}                                                                                                                  \\ \hline
MSCE-BERT & 71.4 & 81.8 & 74.8 & 83.6 & 77.5 & 79.5 & 72.6 & 77.3\\
MCSE-BERT~(reproduced) &   72.3  &  81.3  &  74.5  &  83.7  &  78.7  &     80.2     &      72.0     &  77.5   \\ 
MCSEO-BERT~(ours) &   70.8 &  82.6  &  75.9  &  84.9 &  79.6 &     80.8     &       73.4     &  \textbf{78.3}    \\ \hline
MCSE-RoBERTa &  71.7 & 82.7 &75.9 &84.0 &81.3 &82.3 &70.3 &78.3   \\ 
MCSE-RoBERTa~(reproduced) &   73.8  & 82.2  & 75.1  & 83.0  & 78.7  &    83.0     &     71.5      & 78.2    \\ 
MCSEO-RoBERTa~(ours) &   74.1  & 82.8  & 76.7  & 83.6 & 80.3  &   83.3     &     72.0      & \textbf{79.0}      \\ 

\hline
\end{tabular}}
\label{results}
\caption{Evaluation results on seven STS tasks. We report Spearman's correlation scores for all tasks. Except for our results, all the other results are quoted from their paper.}
\end{table*}
\subsection{Sentence-image level CL}
Given an image-caption pair ${\{d^{img}_i, d^{cap}_i} \}$ from a dataset $\mathcal{D}$, where $i$ stands for the $i_{th}$ example in the dataset, we can use an image encoder $g_\theta$ and a sentence encoder $f_\theta$ to obtain their respective representations: 
\begin{equation}
 \\   h_i^{img} = g_\theta(d^{img}_i),  \ \  h_i^{cap} = f_\theta(d^{cap}_i).
\end{equation} 
Since the caption describes the content of the corresponding image, it is natural to treat the paired image as the positive sample for the caption, while all other images in the batch serve as negative samples.  The training objective for sentence-image level CL can be formulated as~\cite{zhang-etal-2022-mcse}: 
\begin{equation}
  \\  l_{i}^{img-cap} = -\mathrm{log}\frac{e^{sim(h_i^{cap}, h_i^{img})/\tau'}}{\sum_{j=1}^{N} e^{sim(h_i^{cap}, h_j^{img})/\tau'}},
\end{equation}
where $\tau'$ is the temperature parameter.

However, as has been discussed in the previous sections, simply aligning $d_i^{img}$ and $d_i^{cap}$ may introduce noise due to imperfect correspondences. To mitigate this, we propose to further refine the alignment by matching objects in the image with their corresponding phrases in the caption.
\subsection{Extraction of Object-Phrase Pairs}
%Given an image-caption pair ${\{d_{img}^i, d_{cap}^i} \}$ from a dataset $\mathcal{D}$, where $i$ stands for the $i_{th}$ example in the dataset,

We first utilize a detection model ${f_{detect}}$ capable of both segmentation and grounding on input images and captions e.g., SAM2~\cite{ravi2024sam2} combined with Florence2~\cite{florence2}. This allows us to decompose each image–caption pair into detected objects and corresponding phrases:
\begin{equation}
   \\  O_i, P_i = f_{detect}(\{d_{i}^{img}, d_{i}^{cap}\}),
\end{equation}
where $O_i=\{o_i^1, o_i^2, ..., o_i^k\}$ and $P_i=\{p_i^1, p_i^2, ..., p_i^k\}$ represent the sets of detected objects and their matching phrases from the $i_{th}$ image-caption pair. Note that each detected object is guaranteed to have a corresponding phrase, so both sets contain the same number of elements. Specifically, since simply using ${f_{detect}}$ would output bounding boxes and labels, we instead use the detected masks for precisely extracting the objects in the given image.

After that, we utilize a general image encoder to convert the detected objects into image embeddings that have fixed dimensions:
\begin{equation}
    \\    h_o^{i,k} = g_\theta(o_i^{k}), 
\end{equation}
where $o_i^{k}$ is the $k_{th}$ detected object in the $i_{th}$ image. At the same time, we can also convert the decomposed phrases into vectors:
\begin{equation}
    \\    h_p^{i,k} = f_\theta(p_i^{k}),
\end{equation}
where $\texttt{h}_p^{i,k}$ indicates the representation of the $k_{th}$ phrase from the $i_{th}$ caption. During implementation, instead of re-encoding each phrase individually, we obtain phrase embeddings by applying average pooling over the token embeddings of the full sentence at the corresponding phrase positions. This allows us to efficiently extract phrase-level representations while maintaining consistency with the sentence encoder.

\subsection{Object-phrase level CL}
For the $i_{th}$ image-caption pair, we can conduct CL for the $k_{th}$ phrase based on the detected objects and decomposed phrases from this pair:

\begin{equation}
   \\ l_{i}^{obj-phra} = -\displaystyle \sum_{p\in P_i, o\in O_i}\mathrm{log}\frac{e^{sim(h_p^{i,k}, h_o^{i,k})/\tau}}{\sum_{j=1}^{K} e^{sim(h_p^{i,k}, h_o^{i,j})/\tau}},
\end{equation}
where $K$ is the number of elements in set $O_i$ and $P_i$. This formulation encourages each phrase to be most similar to its corresponding object embedding compared to other objects. Note that this object-phrase-level loss is calculated within each image-caption pair instead of the whole batch.

The final loss of MCSEO is the combination of three CL losses:
\begin{equation}
  \\  l_i = l_i^{text} + \alpha l_{i}^{img-cap} + \beta l_{i}^{obj-phra}.
\end{equation}
$\alpha$ and $\beta$ are two hyperparameters that balance the contributions of each objective.

\section{Experiments}
In this section, we begin by introducing implementation details and then discuss our experimental results.

\subsection{Setup}
We conduct experiments based on two kinds of text backbone models: BERT~\cite{devlin-etal-2019-bert} and RoBERTa~\cite{liu2019robertarobustlyoptimizedbert}. Specifically, we choose \texttt{bert-base-uncased} and \texttt{roberta-base}. For the image encoder, we use ResNet50~\cite{resnet} to encode both images and segmented objects into a dimension of 2048. To address the dimensional mismatch between textual embeddings (768 dimensions) and visual embeddings, we introduce a multi-layer perceptron (MLP) projection layer that maps both modalities into a shared embedding space of 256 dimensions. For sentence embeddings, we use the output corresponding to the \texttt{[CLS]} token. The image encoder is kept frozen during training and we only update the text encoder.

Following ~\citet{gao-etal-2021-simcse} and ~\citet{zhang-etal-2022-mcse}, we use the Wikipedia subset as text-only dataset, and Flickr30k~\cite{flickr} as the image-caption dataset. We use cosine similarity as the similarity function in all contrastive objectives, and fix the temperature parameter $\tau$ to 0.05.

To extract object–phrase alignments, we use SAM2 with Florence2\footnote{\url{https://github.com/IDEA-Research/Grounded-SAM-2}} as the detection and grounding model. The detected objects are equipped with transparent backgrounds to avoid potential noise. For model training, we follow MCSE and set the weight of the image-caption contrastive loss $\alpha$ to the optimal value 0.01. The weight $\beta$ for the object-phrase contrastive loss is selected via grid search, with $\beta=0.005$ yielding the best results. All models are trained with a batch size of 64, and hyperparameters are tuned using the development set of STS-B, with evaluation performed every 125 steps.

For evaluation, we select seven standard Semantic Textual Similarity (STS) tasks from the SentEval toolkit~\cite{conneau-kiela-2018-senteval} to assess sentence embedding quality.
To maximize the number of object-phrase pairs extracted from each example, we use the longest caption provided for each image in Flickr30k. This strategy results in an average of 4.2 object-phrase pairs per image. During the training process, since we set the maximum length to 32, some phrases detected from the original caption may be truncated and thus not present in the final input sequence. To ensure the validity of the object-phrase alignment, we exclude such truncated phrases from the object-phrase contrastive loss calculation. Additionally, to avoid degenerate cases, image-caption pairs with only one detected object-phrase pair are also excluded from the object-phrase contrastive learning step.

\subsection{Experimental Results}
The experimental results are presented in Table 1. We compare three categories of models: (1) \textit{text-only} baselines trained solely on Wikipedia, (2) MCSE variants trained on both Wikipedia and Flickr30k, and (3) our proposed method, MCSEO, which leverages both datasets and introduces additional object-phrase-level alignment during training. To ensure a fair comparison with MCSE, we also reproduce MCSE using the longest caption available for each image, consistent with the setup in MCSEO.

From Table~1, we observe several key findings. Firstly, object-phrase contrastive learning boosts performance: incorporating object-phrase-level contrastive learning in MCSEO leads to consistent and notable improvements across all STS tasks, for both BERT and RoBERTa backbones. This highlights the effectiveness of introducing finer-grained alignment signals beyond global image-caption matching.

Notably, both MCSE and MCSEO are trained on the exact same datasets Wikipedia and Flickr30k. The only difference is that MCSEO models additionally capture localized object-phrase alignment. The performance gains observed for MCSEO suggest that the quality of supervision (i.e., finer alignment) plays a crucial role, even when the quantity of data remains unchanged.

Finally, the benefits of MCSEO are observed consistently regardless of the underlying text encoder. This indicates that our object-phrase alignment strategy is model-agnostic and generalizes well across architectures like BERT and RoBERTa.

\section{Conclusion}
In this work, we propose MCSEO, a novel multimodal contrastive learning framework for sentence embeddings that introduces fine-grained supervision by aligning object-phrase pairs extracted from image-caption data. By leveraging existing detection and grounding models, MCSEO decomposes each image and caption into semantically grounded object-phrase pairs and applies an additional object-phrase-level contrastive loss during training.
Extensive evaluations on STS tasks demonstrate that incorporating these finer-grained alignments significantly improves the quality of sentence embeddings across different backbone models. Our results show that MCSEO effectively mitigates the noise inherent in raw image-caption pairs by providing more accurate alignment signals, ultimately leading to better generalization and semantic representation.

\bibliography{anthology,custom,j_yourrefs}
\bibliographystyle{acl_natbib}

% \appendix

% \section{Example Appendix}
% \label{sec:appendix}

% This is a section in the appendix.

\end{document}